\documentclass[sigconf,nonacm]{acmart}

\newcommand{\tabincell}[2]{
\begin{tabular}{@{}#1@{}}#2\end{tabular}
}
\usepackage{array}

\newcolumntype{L}[1]{>{\raggedright\let\newline\\\arraybackslash\hspace{0pt}}m{#1}}
\newcolumntype{C}[1]{>{\centering\let\newline\\\arraybackslash\hspace{0pt}}m{#1}}
\newcolumntype{R}[1]{>{\raggedleft\let\newline\\\arraybackslash\hspace{0pt}}m{#1}}

\usepackage{tabularx}
\usepackage{multirow, graphicx, subfigure,caption,color,hyperref}
\usepackage{arydshln}
\usepackage{makecell}
\usepackage{fancyhdr}
\AtBeginDocument{%
  \providecommand\BibTeX{{%
    \normalfont B\kern-0.5em{\scshape i\kern-0.25em b}\kern-0.8em\TeX}}}

\begin{document}

\title{PicT: A Slim Weakly Supervised Vision Transformer for Pavement Distress Classification}

\author{Wenhao Tang}
\orcid{0000-0002-8835-9215}
\affiliation{%
  \institution{School of Big Data \& Software Engineering, Chongqing University}
  \city{Chongqing}
  \country{China}
  \postcode{400044}
}
\email{whtang@cqu.edu.cn}

\author{Sheng Huang}
\orcid{0000-0001-5610-0826}
\authornote{indicates the corresponding author.}
\affiliation{%
  \institution{School of Big Data \& Software Engineering and Ministry of Education Key Laboratory of Dependable Service Computing in Cyber Physical Society, Chongqing University}
  \city{Chongqing}
  \country{China}
  \postcode{400044}
}
\email{huangsheng@cqu.edu.cn}

\author{Xiaoxian Zhang}
\affiliation{%
  \institution{School of Big Data \& Software Engineering, Chongqing University}
  \city{Chongqing}
  \country{China}
  \postcode{400044}
}
\email{zhangxiaoxian@cqu.edu.cn}

\author{Luwen Huangfu}
\orcid{0000-0003-3926-7901}
\affiliation{%
 \institution{Fowler College of Business and Center for Human Dynamics in the Mobile Age, San Diego State University}
 \city{San Diego}
 \state{California}
 \country{USA}
 \postcode{92182}
 }
 \email{lhuangfu@sdsu.edu}

\renewcommand{\shortauthors}{Accepted by ACM MM '22} 
\begin{abstract}

  Automatic pavement distress classification facilitates improving the efficiency of pavement maintenance and reducing the cost of labor and resources. A recently influential branch of this task divides the pavement image into patches and infers the patch labels for addressing these issues from the perspective of multi-instance learning. However, these methods neglect the correlation between patches and suffer from a low efficiency in the model optimization and inference. As a representative approach of vision Transformer, Swin Transformer is able to address both of these issues. It first provides a succinct and efficient framework for encoding the divided patches as visual tokens, then employs self-attention to model their relations. Built upon Swin Transformer, we present a novel vision Transformer named \textbf{P}avement \textbf{I}mage \textbf{C}lassification \textbf{T}ransformer (\textbf{PicT}) for pavement distress classification. In order to better exploit the discriminative information of pavement images at the patch level, the \textit{Patch Labeling Teacher} is proposed to leverage a teacher model to dynamically generate pseudo labels of patches from image labels during each iteration, and guides the model to learn the discriminative features of patches via patch label inference in a weakly supervised manner. The broad classification head of Swin Transformer may dilute the discriminative features of distressed patches in the feature aggregation step due to the small distressed area ratio of the pavement image. To overcome this drawback, we present a \textit{Patch Refiner} to cluster patches into different groups and only select the highest distress-risk group to yield a slim head for the final image classification. We evaluate our method on a large-scale bituminous pavement distress dataset named CQU-BPDD. Extensive results demonstrate the superiority of our method over baselines and also show that \textbf{PicT} outperforms the second-best performed model by a large margin of $+2.4\%$ in P@R on detection task, $+3.9\%$ in $F1$ on recognition task, and 1.8x throughput, while enjoying 7x faster training speed using the same computing resources. Our codes and models have been released on \href{https://github.com/DearCaat/PicT}{https://github.com/DearCaat/PicT}.
\end{abstract}

  \ccsdesc[500]{Computing methodologies~Object identification}
  \ccsdesc[300]{Computing methodologies~Computer vision tasks}
  \ccsdesc[500]{Computing methodologies~Object recognition}

\keywords{Pavement Distress Classification, Deep Learning, Image Classification, Transformers, Weakly Supervised Learning}

\maketitle

\section{Introduction}
\label{sec:intro}



With the rapid growth of transport infrastructures such as airports, bridges, and roads, pavement maintenance is deemed as a crucial element of sustainable pavement today~\cite{pavement_maintain}. It is a trend to automate this process via machine learning and pattern recognition techniques, which can significantly reduce the cost of labor and resources. One of the core tasks in pavement maintenance is pavement distress classification (PDC), which aims to detect the damaged pavement and recognize its specific distress category. These two steps are also referred to as pavement distress detection and pavement distress recognition, respectively~\cite{wsplin_trans}.


\begin{figure}[t]
    \centering
    \includegraphics[width=8.5cm]{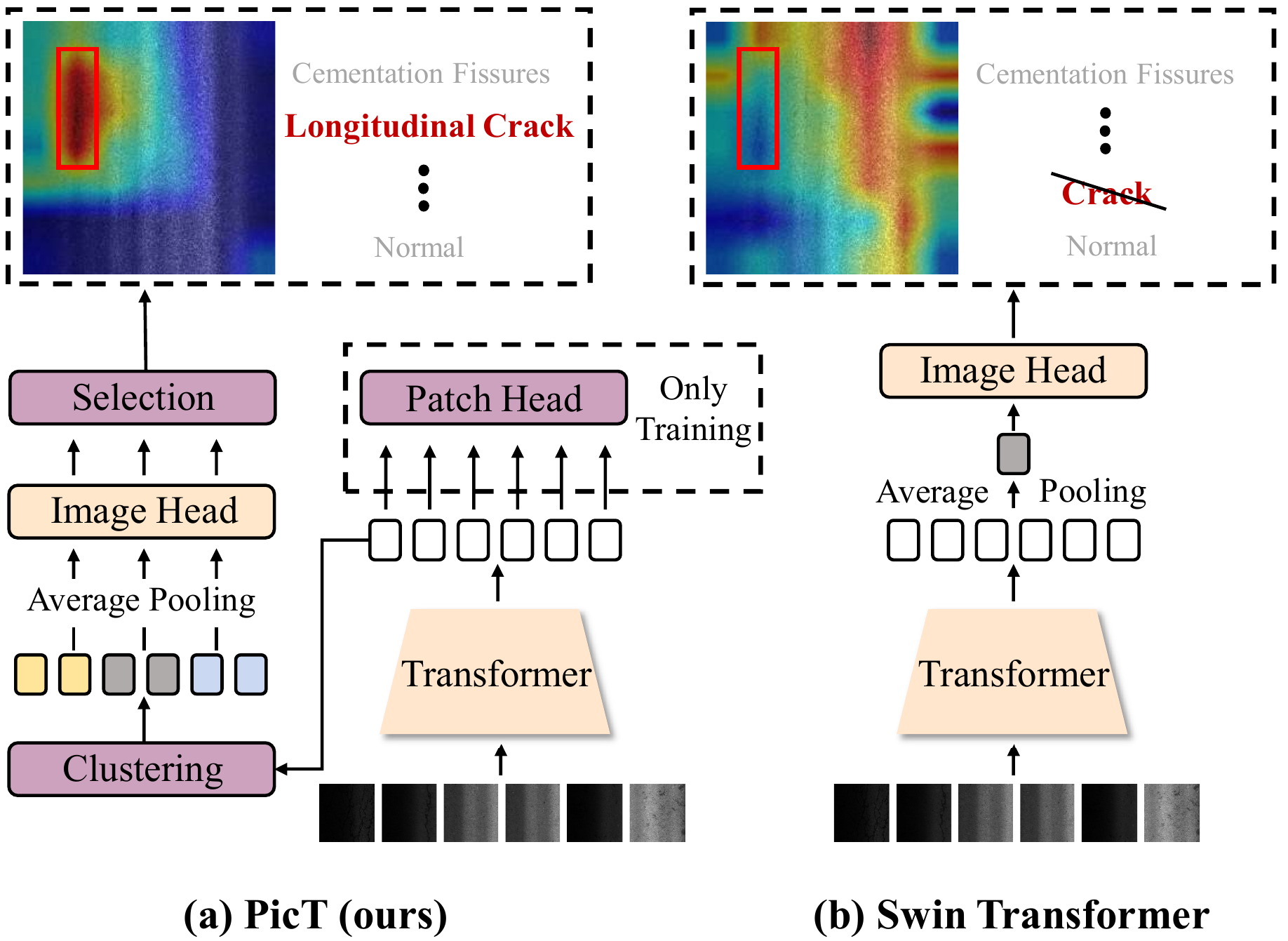}  \vspace{-0.65cm}
    \caption{(a) The proposed PicT aims to learn the discriminative information of patches via performing the patch classification in a weakly supervised manner while employing a slim image classification head to select the features of the highest risk patches only for achieving a better distress classification. (b) In contrast, the general vision Transformer, like Swin Transformer, fails to mine the local discriminative features, and locate the distressed area. Moreover, its classifier is easy to be fooled by the overabundant features of non-distressed patches due to the small distressed area ratio property of pavement images.}
    \label{fig.first_page}
    \vspace*{-0.7cm}
\end{figure}


For decades, many researchers have been devoted to investigate PDC from the perspective of computer vision. The traditional methods are to utilize image processing, hand-craft features, and conventional classifiers to recognize the pavement distress~\cite{c3,c28,c29,c30}. The main drawback of these methods is that they highly rely on expert knowledge and suffer from the weak generalization ability~\cite{dong2021automatic}. Inspired by the recent remarkable successes of deep learning, many deep learning-based approaches have been proposed for addressing the PDC issue~\cite{c8,c10,c11}. However, these methods often regard the PDC problem as a common image classification problem, while seldom paying attention to the specific characteristics of pavement images, such as the high image resolution, the low distressed area ratio, and uneven illumination, in the model design phase~\cite{wsplin}.

\begin{figure}[t]
  \centering
  \includegraphics[width=8.5cm]{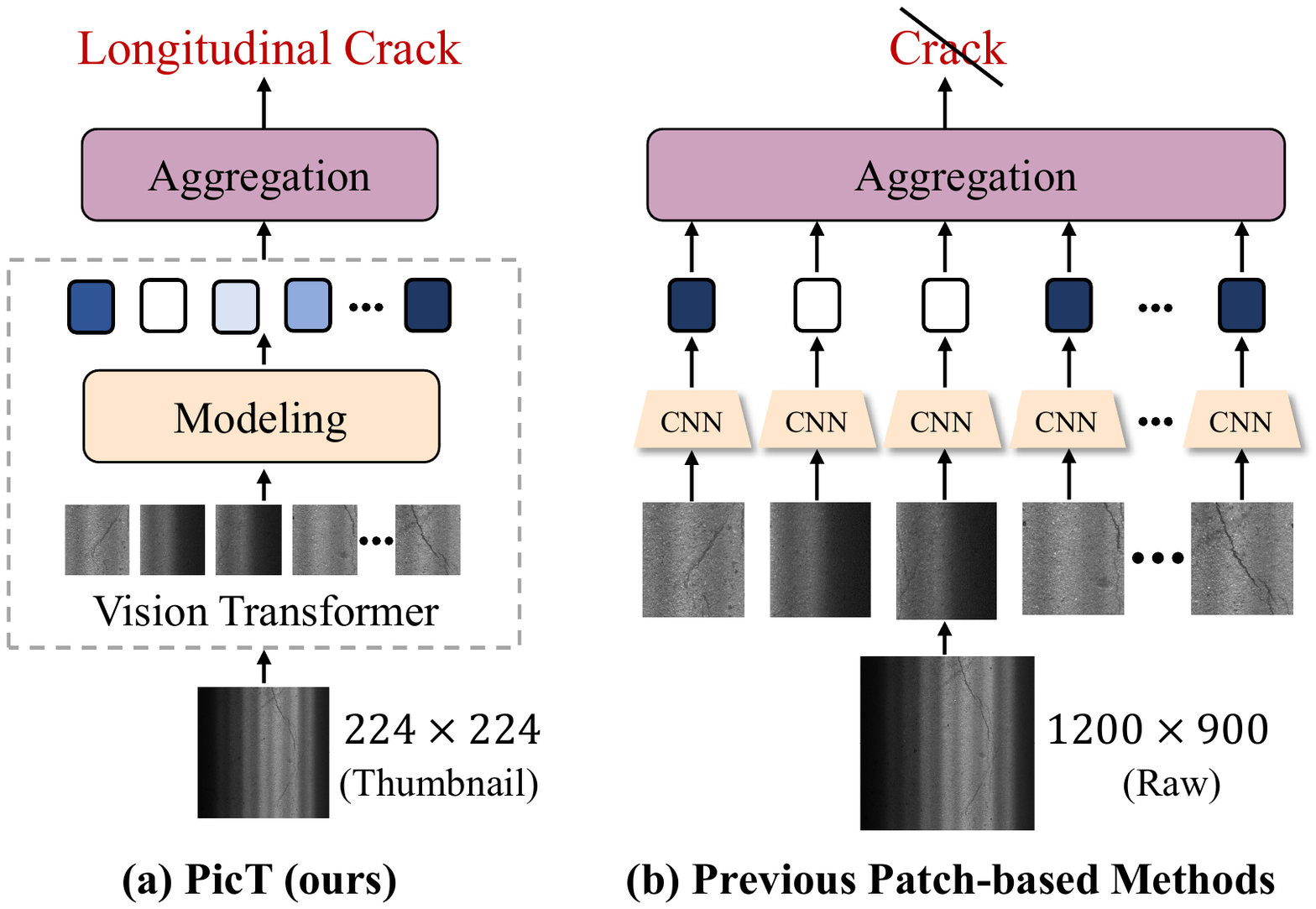}  \vspace{-0.6cm}
  \caption{ (a) The PicT is a more succinct and efficient learning framework based on the vision Transformer and leverages the self-attention module to model the correlation between patches. (b) In contrast, because CNN lacks the ideal inter-local correlation modeling capability and struggles to capture the subtle discriminative features, previous patch-based methods have to manually split the high-resolution raw image into patches. This leads to lower efficiency and the neglect of the patch correlation.}
  \label{fig.contrs}
  \vspace{-0.6cm}
\end{figure}

To consider these characteristics of pavement images, some of the latest approaches~\cite{ioplin,wsplin,wsplin_trans} attempt to divide the high-resolution raw pavement image into patches and solve this problem via inferring the labels of patches with CNNs as shown in Figure~\ref{fig.contrs}. Intuitively, these approaches can be deemed as a kind of multi-instance learning approach~\cite{dietterich1997mil}, merited by their ability to better learn the local discriminative information of the high-resolution image and thereby achieve promising performances. For example, IOPLIN~\cite{ioplin} conducts the feature extraction and classification on each patch independently and finally aggregates the classification results via max-pooling or average-pooling. Similarly, WSPLIN~\cite{wsplin,wsplin_trans} also extracts the feature of each patch and classifies the patch independently. However, instead of conducting the classification result aggregation, it inputs the classification results of all patches into a comprehensive decision network for accomplishing the final classification. Generally speaking, the whole patch label inference process, including patch partition, feature extraction, and classification is quite expatiatory, which leads to the low efficiencies of these approaches. What's more, the patches naturally have correlation while these methods all neglect this nature.

Recently, the vision Transformer has demonstrated excellent performances in supervised learning~\cite{shao2021transmil,video_class,xie2021segformer,girdhar2019video}, weakly supervised learning~\cite{simeoni2021token_local,gao2021trans_cam}, and self-supervised learning tasks~\cite{xie2021simmim,he2021mae}. Its success is benefited from the self-attention mechanism~\cite{attention} and the strong local representation capability. As an influential approach in vision Transformer, Swin Transformer~\cite{swin} shares a similar idea as the aforementioned patch-based PDC in the visual information exploitation. It also partitions the image into different patches and encodes them as visual tokens. Then, self-attention is employed to model their relations. Finally, these tokens are aggregated for yielding the final classification. Compared to previous approaches, the vision Transformer provides a more succinct and advanced framework for learning the subtle discriminative features of images as shown in Figure~\ref{fig.contrs} (a). However, it is still inappropriate to directly apply Swin Transformer to address the PDC issue for two reasons. The first one is that the potential discriminating features of patches are still not well explored as the CAMs shown in Figure~\ref{fig.first_page} (b), since the patch-level supervised information has still not been sufficiently mined in the Swin Transformer. The second reason is that its average-pooling aggregation for all patches will suppress the distressed patch features, since the distressed patches are only a small fraction of all patches.

In this paper, we elaborate on a novel Transformer named \textbf{PicT} which stands for \textbf{P}avement \textbf{I}mage \textbf{C}lassification \textbf{T}ransformer, based on the Swin Transformer framework for Pavement Distress Classification (PDC). In order to overcome the first drawback of the Swin Transformer in PDC, we develop a \textit{Patch Labeling Teacher} module based on the idea of teacher-student model to present a weakly supervised patch label inference scheme for fully mining the discriminative features of patches. In order to overcome the second drawback of the Swin Transformer in PDC, a \textit{Patch Refiner} module is designed to cluster the patches and select the patches from the highest risk cluster to yield a slimmer head for final classification. This strategy can significantly suppress the interferences from the plentiful non-distressed (normal) patches in the image label inference phase. A large-scale pavement image dataset named CQU-BPDD is adopted for evaluation. Extensive experiments demonstrate that PicT achieves state-of-the-art performances in two common PDC tasks, namely pavement distress detection, and pavement distress recognition. More importantly, PicT outperforms the second-best performed model by a large margin of +2.4\% in P@R on detection task, +3.9\% in $F1$ on recognition task, and 1.8x throughput, while enjoying 7x faster training efficiency.

The main contributions of our work are summarized as follows:
\vspace{-0.45cm}
\begin{itemize}
    \item We propose a novel vision Transformer named PicT for pavement distress classification. PicT not only inherits the merits of Swin Transformer but also takes the characteristics of pavement images into consideration. To the best of our knowledge, it is the first attempt to specifically design a Transformer to address pavement image analysis issues.
    \item A \textit{Patch Labeling Teacher} module is elaborated to learn to infer the labels of patches in a weakly supervised fashion. It enables better exploiting the discriminative information of patches under the guidance of the patch-level supervised information, which is generated by a \textit{Prior-based Patch Pseudo Label Generator} based on image labels. Such an idea can also be flexibly applied to other vision Transformers to infer labels of visual tokens in a weakly supervised manner.
    \item A \textit{Patch Refiner} is designed to yield a slimmer head for PDC. It enables filtering out the features from low-risk patches while preserving the ones from high-risk patches to boost the discriminating power of the model further.
    \item We systematically evaluate the performances of common vision Transformers on two pavement distress classification tasks. Extensive results demonstrate the superiority of PicT over them. Compared to the second-best performed method, PicT achieves 2.4\% more detection performance gains in P@R, 3.9\% recognition performance gains in $F1$, 1.8x higher throughput, and 7x faster training speed.
\end{itemize}
\vspace{-0.35cm}

\section{Related Work}
\label{sec:rela_work}

\subsection{Pavement Distress Classification}
Pavement Distressed Classification (PDC) aims to detect the diseased pavement and recognize the specific distress category of pavements with pavement images. The traditional methods are to utilize image processing, hand-craft features, and conventional classifiers to recognize the pavement distress~\cite{c3,c28,c29,c30}. The main drawback of these methods is that they rely on expert knowledge and suffer from the weak generalization ability~\cite{dong2021automatic}.

Inspired by the recent remarkable successes of deep learning in extensive applications, simple and efficient convolutional neural networks (CNN) based PDC methods have gradually become the mainstream in recent years. There are two major branches: one is the general-CNN methods and the other is patch-based methods. The general-CNN approaches~\cite{c8,c10,c11} only regard the PDC as a standard image classification problem and directly apply the classical deep learning approaches to solve it. For example, ~\cite{c10} proposes a novel method using deep CNN to automatically classify image patches cropped from 3D pavement images and successfully trains four supervised CNNs with different sizes of receptive field. However, general-CNN methods seldom paid attention to the specific characteristics of pavement images. Patch-based methods~\cite{ioplin,wsplin_trans,wsplin} have addressed these issues by splitting the patches and inferring the labels of patches. IOPLIN~\cite{ioplin} manually partitions the high-resolution pavement image into patches and elaborates an iteratively optimized CNN model to predict patch labels for detection. It enables learning the discriminative details of the local regions and then improves the performance. WSPLIN~\cite{wsplin,wsplin_trans} conducts the same process as IOPLIN but uses a comprehensive decision network to aggregate the classification results of patches for accomplishing the final image classification. However, the whole image partition, patch feature extraction, and classification process often necessitates a large burden in efficiency. Moreover, such methods often neglect the correlation between patches.
\subsection{Vision Transformer}
Transformer origins from natural language processing and machine translation~\cite{attention,devlin2018bert}. Recently, they have been successfully applied to many computer vision domains~\cite{li2021video,chen2021dpt,zhang2021reid,video_class,face,he2021transrefer3d}. Generally speaking, the Transformer models in the computer vision community can be roughly divided into two groups: combining self-attention with CNN~\cite{girdhar2019video,xie2021segformer,shao2021transmil,feng2021convolutional} and vision Transformer models~\cite{vit,swin,xie2021simmim,xu2021soft}. The first type of method focuses on using powerful self-attention mechanism to model the correlation between features extracted by CNN. For example, Girdhar et al.~\cite{girdhar2019video} repurpose a Transformer-style architecture to model features from the spatiotemporal context around the person whose actions are attempted for classification. On the other hand, with the release of ViT~\cite{vit}, more and more work apply vision Transformer to achieve better results in different vision areas~\cite{swin,xu2021soft,gao2021trans_cam,xie2021simmim,he2021mae,bao2021beit,transfg,hu2021fgia}. For example, LOST~\cite{simeoni2021token_local} achieves SOTA on objective search by combining similar tokens and using seed search to complete the object localization. SimMM~\cite{xie2021simmim} predicts raw pixel values of the randomly masked patches by a lightweight one-layer head and performs learning using a simple L1 loss. In particular, these outstanding studies in the field of weakly supervised localization~\cite{simeoni2021token_local,gao2021trans_cam} and self-supervised learning~\cite{xie2021simmim,he2021mae,bao2021beit} demonstrate the powerful local representation capabilities of tokens in vision Transformer models. In this work, we attempt to leverage the merits of Transformer for addressing the PDC issue.

\begin{figure*}[t]
  \centering
  \includegraphics[width=13cm]{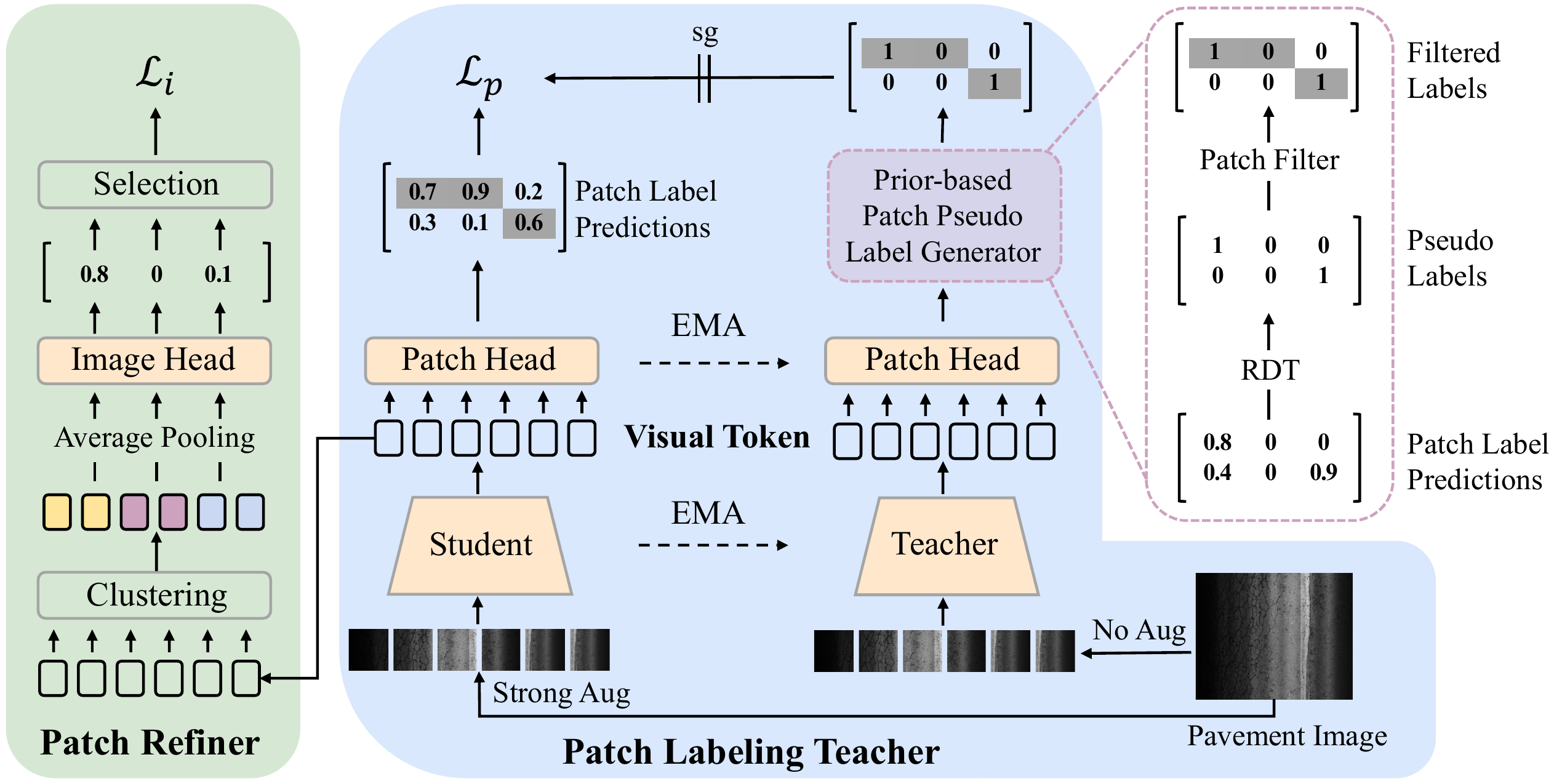}
  \caption{The overview of the PicT model. PicT involves two classification branches in the training phase. One is the \textit{Patch Labeling Teacher}, which implements classification at the patch level. While the other is the \textit{Patch Refiner}, which conducts the classification at the image level. "sg" indicates stopping the gradient of teacher model.}
  \label{fig.network}
  \vspace{-0.2cm}
\end{figure*}

\section{Method}
\label{sec:format}



\subsection{Problem Formulation and Overview}
\label{sec:formulation}
\subsubsection{\textbf{Problem Formulation}}  Pavement Distress Classification (PDC) can be deemed as an image classification task from the perspective of computer vision. Let $X=\{x_1,\cdots,x_n\}$ and $Y=\{y_1,\cdots,y_n\}$ be the collection of pavement images and their pavement labels respectively, where $y_i$ is a $ C $-dimensional vector and $C$ is the number of categories. Let the subset $Y_{nor} \in Y$ be the non-distressed label set and $Y_{dis} \in Y$ be the distressed label set, which indicate the presence or absence of distress in a pavement image. Pavement distress detection and recognition are the most common tasks in the PDC. The pavement distress detection task is a binary image classification issue ($C=2$.), which judges if a pavement image exists distress or not, while the pavement distress recognition task is a multi-class image classification problem ($C>2$) which classifies the pavement image into a specific distress category. In this paper, we develop a novel vision Transformer model named PicT for addressing the aforementioned PDC issues both.

\vspace{-0.29cm}
\subsubsection{\textbf{Overview}} The overview of PicT is shown in Figure~\ref{fig.network}. This entire framework is based on the well-known Swin Transformer. In order to incorporate the specific characteristics of pavement images, such as the high resolution and the low distressed area ratio, we enhance Swin Transformer by introducing two elaborate modifications. We borrow the idea from self-supervised learning to develop a \textit{Patch Labeling Teacher} (PLT) for better exploring the discriminating power of patch information via classifying patches in a weakly supervised manner. The main obstacle to patch classification is that there are the only image-level labels for training. So, we design a \textit{Prior-based Patch Pseudo Label Generator} (P$^3$LG) for dynamically generating the pseudo labels of patches based on the prior information of image labels and distressed area ratio during the optimization of PLT. Since only a small fraction of patches has distresses, the original global-average-pooling based feature aggregation easily leads to the problem that the discriminative features of the distressed patches are diluted by the features from the overabundant non-distressed patches. In order to solve this problem, the \textit{Patch Refiner} (PR) is presented to cluster patches into different groups, and then only the patches from the highest distressed risk group are aggregated to yield a slim head for image classification. In the following sections, we will introduce these modules in detail.

\vspace{-0.2cm}
\subsection{Patches Labeling Teacher}
\label{sec:pt}

Similar to the Swin Transformer, PicT splits the resized image into patches first, and employs the Swin Transformer blocks as the feature extractor to learn the visual features of patches, which are also referred to as tokens.
The Swin Transformer lacks the patch-level supervision, which leads to the insufficient exploration of discriminative information in patches. We present a \textit{Patch Labeling Teacher} (PLT) module to address this issue. Similar to the idea of self-supervised learning, PLT also adopts the teacher-student scheme, which employs a teacher model to supervise the optimization of the student model (the finally adopted model). More specifically, let $\mathcal{F}_{\theta_{s}  }\left ( \cdot  \right )$ and $\mathcal{F}_{\theta_{t} }\left ( \cdot  \right )$ be the mapping functions of the student and teacher models respectively. $\theta_s$ and $\theta_t$ are their corresponding parameters. The parameters of the teacher model $\theta_{t} $ are updated by the exponential moving average (EMA) of the student parameters $\theta_{s} $. The update rule is $ \theta_{t} \leftarrow \lambda \theta_{t} + \left( 1-\lambda\right) \theta_{s}$. The student and teacher models share the same network structure. They all use Swin-S~\cite{swin} as the backbone $f\left ( \cdot  \right )$, and a patch head $h^{pat}\left ( \cdot  \right ) $ for classifying the patches, $\mathcal{F}  = h^{pat}\circ f$. The patch head only consists of a Linear layer with $L$ dimensions where $L$ is the dimension of token. The Swin-S architecture considers pavement image as a grid of non-overlapping contiguous image patches, and these patches are then passed through Swin Transformer blocks to form a set of features, which are called tokens. The token collection can be denoted as,
\begin{equation}
T_{i} = f(x_{i}) = \left \{ t_{i1},\dots ,t_{ij},\dots,t_{im}  \mid  t_{ij}\in \mathbb{R}^{N \times N \times L} \right\},
\end{equation}
where $t_{ij}$ is the $j$-th token of the $i$-th image $x_i$, $m=49,N=16$ and $L=768$. The patch label predictions can be achieved by inputting these tokens of image $x_i$ to the patch head for classification,
\begin{equation}
  P_{i} = h^{pat}(T_{i})=\left \{ p_{i1},\dots ,p_{ij},\dots,p_{im}  \mid p_{ij}\in \mathbb{R}^{C}\right \},
\end{equation}
where $p_{ij}$ is the label prediction of token $t_{ij}$. However, we only have the image-level label while do not have patch-level label for training the patch classifier directly. In such a manner, we design a \textit{Prior-based Patch Pseudo Label Generator} (P$^3$LG) to generate the pseudo labels of patches (tokens) based on the image-level label $y_i$ and the label predictions produced by the teacher model in the previous iteration, which will be introduced later. The pseudo patch labels of image $x_i$ are denoted as $\tilde{Y}_i=\{\tilde{y}_{i1},\cdots,\tilde{y}_{ij},\cdots,\tilde{y}_{im}\}$, where $\tilde{y}_{ij}$ is the pseudo label of patch $t_{ij}$.

Finally, we use the cross-entropy to measure the discrepancy between the patch pseudo label and the student patch label prediction, and denote the patch classification loss as follows,
\begin{equation}
  \mathcal{L}_{p} = -\frac{1}{nm}\sum_{i=1}^{n}\sum_{j=1}^{m}\tilde{y}_{ij}\log p_{ij}.
\end{equation}

Moreover, to facilitate the training of the student model, we adopt FixMatch~\cite{sohn2020fixmatch}, which is a popular advancement in the semi-supervised image classification task. Strong augmented data is applied for training the student model while the original data is used for training the teacher model.

\subsection{Prior-based Patch Pseudo Label Generator}
\label{sec:ppl}


The performance of the patch classifier highly relies on the quality of the generated pseudo-label~\cite{xu2021soft}. So, we propose a \textit{Prior-based Patch Pseudo Label Generator} (P$^3$LG) to accomplish this task with the help of the prior information, such as image labels, distressed area ratio, and the patch label predictions dynamically produced by the teacher model. P$^3$LG consists of two important steps: Relative Distress Threshold and Patch Filter.

\subsubsection{\textbf{Relative Distress Threshold (RDT)}}
A Relative Distress Threshold (RDT) is defined for producing the pseudo labels of patches. In RDT, the patch pseudo labels are always fixed to the normal one, $y_{normal}$, if the pavement image has no distress ($y_{i} \in Y_{nor}$). While the patches, which own the top $\Delta _{rel}\%$ highest predictions of the image distress label $y_i$, will be considered as the patches owning the distressed category $y_i$, if the distressed label of the pavement image is $y_i$. The whole strategy can be mathematically denoted as follows,
\begin{equation}\label{thr_relative}
  \tilde{y}_{ij} =  \begin{cases}
    y_{i}, & {p^{y_{i}}_{ij}> \mathcal{T}\left ( \Delta _{rel} \right ) ~\mathrm{and}~ y_i \in Y_{dis}}\\
    y_{normal}, & others,
  \end{cases}
\end{equation}
where $p^{y_{i}}_{ij}$ is the element of patch label predictions corresponding to the distressed category of $y_i$, and $\mathcal{T} \left( \Delta _{rel} \right)$ returns the $\Delta _{rel}\% $ largest $p^{y_{i}}_{ij}$ of all patches in image $x_i$. Compared with the common absolute threshold strategy~\cite{abl_thr}, this threshold enables ensuring that a distressed image always has a certain percentage of distressed instances. This also alleviates the pseudo-label bias caused by inconsistent training of the model for different distress categories.

\subsubsection{\textbf{Patch Filter}} Following~\cite{xu2021soft}, we filter out low prediction confidence patches, and only preserve the high confidence ones for training.
\begin{itemize}
\item For \textbf{Distressed Image}: We preserve all patches whose pseudo-labels are not $y_{normal}$, and also preserve the patches whose pseudo-labels are $y_{normal}$ and $ p_{ij}^{y_{normal}}>0.5$.
\item For \textbf{Non-distressed Image}: We only preserve the high prediction confident patches whose  $p_{ij}^{y_{normal}}>0.95$.
\end{itemize}
These two optimal patch filtering thresholds are obtained in an empirical way. 

\subsection{Patch Refiner}
\label{sec:pc}
After the feature extraction of patches, these features will be aggregated via average-pooling overall patches, and then performed for classification in the Swin Transformer. However, the distressed patches are only a small fraction of overall patches, so this feature aggregation strategy is easy to dilute the discriminative features of distressed patches. In order to address this issue, we develop a \textit{Patch Refiner} (PR) to yield a slimmer image head.

Let $\mathcal{R}_{\xi }\left( \cdot\right)$ be the mapping function of PR where $\xi$ is its parameters. It consists of three steps, namely token clustering, aggregation, and selection. The mapping functions of these steps are denoted as $f_{clu}(\cdot)$, $h^{img}(\cdot)$, and $f_{sel}(\cdot)$ respectively. So $\mathcal{R} = f_{sel}\circ h^{img}\circ f_{clu}$. Compared with the image head in Swin Transformer, which aggregates all tokens for classification, our image head is a much slimmer head, since it only highlights the highest risk group of tokens for conducting the final classification.



In \textit{Patch Refiner}, the tokens are first clustered into different groups via using K-Means~\cite{kmeans}. Then the average-pooling is conducted on each group for feature aggregation, and these aggregated features will be input into an image head for assessing the distressed risk of each group. The whole process can be mathematically represented as follows,
\begin{equation}
  R_{i} = h^{img} \left(f_{clu} \left( T_{i} \right)\right)=\left\{r_{i1},\cdots,r_{it},\cdots,r_{ik}\mid r_{it}\in \mathbb{R}^{C}\right\},
\end{equation}
where $R_{i}$ is the label predictions of groups, and $k$ is the group number. $r_{it}$ is the label predictions of the $t$-th group, which can be deemed as the risk assessments of different distresses for this group. In our approach, $k$ is empirically fixed to $2$ for the detection task while $3$ for the recognition task.

In the image label classification phase, we intend to suppress the interferences from the overabundant non-distressed patches. So, we only select the highest risk group to represent the pavement image for final classification. In other words, we only preserve the label predictions of the group, which owns the lowest confident label prediction in the normal (non-distressed) category, as the predicted labels of image $\hat{y}_{i} $,
\begin{equation}
    \hat{y}_{i} = \mathcal{R} \left( T_{i} \right) = f_{sel}\left( R_{i} \right)={\arg\underset{r_{it}}\min}~~\left\{r_{it}^{y_{normal}} \mid r_{it} \in R_i \right \},
\end{equation}
where $r_{it}^{y_{normal}}$ is the element of $r_{it}$ corresponding to the normal category. Once we obtain the predicted labels of images, we can use the cross-entropy to measure the image classification loss,
\begin{equation}
    \label{eq:cel}
    \mathcal{L}_i = -\frac{1}{n}\sum_{i=1}^{n} y_{i}\log{\hat{y}_i}.
\end{equation}
Finally, the optimal PicT model can be learned by minimizing both the patch and image classification losses,
\begin{equation}
  (\hat{\theta_{s}} ,\hat{\xi}  )  \leftarrow\arg\underset{\theta_{s} ,\xi }\min~~\mathcal{L}_{total}:=\mathcal{L}_i+\mathcal{L}_p.
\end{equation}


\subsection{Pavement Distress Classification}
\label{sec:pavement DR}
 To tackle pavement distress classification, we can train our model as a pavement image classifier. Once the model is trained, the pavement image $x_{i}$ can be fed into PicT for yielding the final classification result. Of particular interest, although we use patch-level branch named \textit{Patch Labeling Teacher} to enhance the use of patch information in the training process. But in the testing phase, we only leverage the image-level branch named \textit{Patch Refiner} for model inference due to the instability of patch-level testing. The process of PDC can be denoted as:
\begin{equation}
    y_{i} = \mathcal{R}\left( f \left( x_{i}\right)  \right)  = f_{sel}\left( h^{img}\left( f_{clu}\left(  f\left(x_{i}\right) \right)  \right)  \right) .
\end{equation}
Note, the selection $f_{sel}\left(\cdot\right)$ in the testing phase is slightly different from it in training phase. More details about this will be specified in the support material.



\section{Experiments}
\subsection{Dataset and Setup}
\subsubsection{\textbf{Task Settings}}
Following~\cite{wsplin_trans}, we evaluate our method on pavement distress detection and recognition tasks under two application settings. The first one is the one-stage detection (\textbf{I-DET}), which is the conventional detection fashion. In this setting, the model only needs to determine whether there are distressed areas in the pavement image. The second one is the one-stage recognition (\textbf{I-REC}), which tackles the pavement distress detection and recognition tasks jointly. In this setting, the model not only detects the presence of distressed areas within the pavement image, but also needs to further classify the distressed category. Compared to \textbf{I-DET}, this task is more challenging and practically applicable. Moreover, both the detection and recognition performances can be evaluated in this setting.

\subsubsection{\textbf{Datasets}}
Before ~\cite{ioplin}, most of the methods in the PDC are validated on private datasets. Thus, following~\cite{ioplin}, a large-scale public bituminous pavement distress dataset named CQU-BPDD~\cite{ioplin} is used for evaluation. This dataset involves seven different types of distress and non-distressed categories. There are 10137 images in the training set and 49919 images in the test set only with the image-level labels. For the setting of \textbf{I-DET}, only the binary coarse-grained category label (distressed or normal) is available. For the setting of \textbf{I-REC}, their fine-grained category labels are available for training and testing models.

\subsubsection{\textbf{Evaluation Protocols and Metrics}}
Following the standard evaluation protocol in~\cite{ioplin}, for \textbf{I-DET} task, we adopt Area Under Curve (AUC) of Receiver Operating Characteristic (ROC) and Precision@Recall (P@R) to evaluate the model performance. The AUC is the common metric in binary classification task. P@R is used to discuss the precision under high recall, which is more meaningful in the medical or pavement distress classification tasks. Since the miss of the positive samples (the distressed sample) may lead to a more severe impact than the miss of the negative ones. As for \textbf{I-REC} task, following~\cite{wsplin_trans}, we use the Top-1 accuracy and Marco F1 score ($F1$) to evaluate the performance of the models. Please refer
to Supplementary for the details of implementation.

\begin{table}[tb]
  \small
  \caption{The pavement distress detection performances of different approaches on the CQU-BPDD. P@R = $n$\% indicates the precision when the corresponding recall is equal to $n$\%. The best performances are in bold, and the second-best results are underlined. }
  \vspace{-0.3cm}
  \centering
  \begin{tabularx}{8.5cm}{l X<{\centering} X<{\centering}X<{\centering}}
  \toprule
   Detectors(\textbf{I-DET}) & AUC & P@R=90\% & P@R=95\% \\
\midrule
  ResNet-50~\cite{res}  & 90.5\% & 45.0\% & 35.3\% \\[1.5pt]
  Inception-v3~\cite{inception} & 93.3\% & 56.0\% & 42.3\% \\[1.5pt]
  VGG-19~\cite{vgg}  & 94.2\% & 60.0\% & 45.0\% \\[1.5pt]
  EfficientNet-B3~\cite{effi}& 95.4\% & 68.9\% & 51.1\% \\[1.5pt]
  ViT-S/16~\cite{vit}  & 95.4\% & 67.7\% & 51.0\% \\[1.5pt]
  ViT-B/16~\cite{vit}  & 96.1\% & 71.2\% & 56.1\% \\[1.5pt]
  DeiT-S~\cite{touvron2021deit} & 95.5\% & 68.2\% & 52.2\% \\[1.5pt]
  DeiT-B~\cite{touvron2021deit} & 96.3\% & 75.2\% & 59.3\% \\[1.5pt]
  Swin-S~\cite{swin} & 97.1\% & 79.9\% & 65.5\% \\[1.5pt]
  IOPLIN~\cite{ioplin} & 97.4\% & 81.7\% & 67.0\% \\[1.5pt]
  WSPLIN-IP~\cite{wsplin_trans} & \underline{97.5\%} & \underline{83.2\%} & \underline{69.5\%} \\[1.5pt]
  {\bf PicT (ours)} & {\bf 97.9\%} & {\bf 85.6\%} & {\bf 75.2\%}\\[1.5pt]
   \midrule
  Recognizers(\textbf{I-REC}) & AUC & P@R=90\% & P@R=95\% \\[2pt]
    \midrule
  EfficientNet-B3~\cite{effi}& 96.0\% & 77.3\% & 59.9\% \\[1.5pt]
  Swin-S~\cite{swin}  & 97.2\% & 82.7\% & 69.0\% \\[1.5pt]
  WSPLIN-IP~\cite{wsplin_trans} & \underline{97.6\%} & \underline{85.3\%} & \underline{72.6\%} \\[1.5pt]
  {\bf PicT (ours)} & {\bf 98.1\%} & {\bf 87.6\%} & {\bf 77.0\%} \\[1.5pt]
  \bottomrule
  \end{tabularx}
  \label{tab:compare_binary}
  \vspace{-0.6cm}
\end{table}

\subsection{Pavement Distress Detection}\label{pdd}
Table~\ref{tab:compare_binary} tabulates the pavement distress detection performances of different methods on \textbf{I-DET} and \textbf{I-REC} settings.
The observations demonstrate that PicT outperforms all the compared methods on both \textbf{I-DET} and \textbf{I-REC} settings under all evaluation metrics. Swin-S is adopted as the backbone of PicT. The performance gains of PicT over it are 0.8\%, 5.7\%, and 9.7\% in AUC, P@R=90\%, and P@R=95\% respectively in the \textbf{I-DET} setting. These numbers on \textbf{I-REC} settings are 0.9\%, 4.9\%, and 8.0\% respectively. These results validate the effectiveness of our modification in Swin-S according to the characteristics of the pavement images.
Moreover, even though Swin-S adopts a more powerful learning framework (Transformer) and also can be deemed as a patch-based approach, it still cannot defeat the conventional patch-based methods, such as WSPLIN and IOPLIN. This phenomenon reflects that it is inappropriate to directly apply Swin-S for PDC, and further confirms that it is important to incorporate the characteristics of pavement images in the model design phase.

WSPLIN is the second-best approach for pavement distress detection. Our method still gets considerable advantages in performance over it. For example, the performance gains of PicT over WSPLIN are 0.4\%, 2.4\%, and 5.7\% in AUC, P@R=90\%, and P@R=95\% respectively on \textbf{I-DET} setting. In the \textbf{I-REC} setting, these numbers are 0.5\%, 2.3\%, and 4.4\% respectively. These results verify the advantages of the Swin Transformer framework over the conventional CNN-based learning framework in pavement distress detection. 

\begin{table}[tb]
  \small
  \centering
  \caption{The pavement distress recognition performances of different methods. $F1$ indicates the Marco F1-score.}
  \vspace{-0.3cm}
  \begin{tabularx}{8cm}{p{3.3cm}X<{\centering}X<{\centering} }
  \toprule
  Recognizers(\textbf{I-REC})  & Top-1& $F1$\\\midrule
ResNet-50 \cite{res}& 88.3\%  & 60.2\%    \\[1.5pt]
VGG-16 \cite{vgg}& 87.7\%  & 58.4\%      \\[1.5pt]
Inception-v3 \cite{inception}& 89.3\%  & 62.9\%\\[1.5pt]
EfficientNet-B3~\cite{effi}& 88.1\% &63.2\%\\[1.5pt]
ViT-S/16~\cite{vit}& 86.8\%  & 59.0\%    \\[1.5pt]
ViT-B/16~\cite{vit}& 88.1\%  & 61.2\%    \\[1.5pt]
DeiT-S~\cite{touvron2021deit}& 88.3\%  & 60.7\%    \\[1.5pt]
DeiT-B~\cite{touvron2021deit}& 89.1\%  & 62.7\%    \\[1.5pt]
Swin-S~\cite{swin}& 89.5\%  & 64.9\%    \\[1.5pt]
WSPLIN-IP~\cite{wsplin_trans} & \underline{91.1\%} &\underline{66.3\%}\\[1.5pt]
\textbf{PicT (ours)}& \textbf{92.2\%} &\textbf{70.2\%}\\[1.5pt] \midrule
  \end{tabularx}
\label{rec}
\vspace{-0.6cm}
\end{table}

\subsection{Pavement Distress Recognition}
Table~\ref{rec} reports the pavement distress recognition performances on the CQU-BPDD dataset. We can observe similar phenomena as the ones observed in pavement distress detection. PicT still outperforms all approaches in pavement distress recognition, which is considered a more challenging PDC task than pavement distress detection. The results show that our method even holds the more prominent advantages in performance compared with the ones of pavement distress detection. For example, the performance gains of PicT over Swin-S are 2.7\% and 5.3\% in Top-1 and $F1$, respectively, while the number of PicT over WSPLIN are 1.1\% and 3.9\% respectively. Clearly, all observations well verify the arguments we raised in Section~\ref{pdd}.

\subsection{Efficiency Evaluation}
We tabulate the training and inference efficiencies of the patch-based approaches and also their backbone networks along with their classification performances in Table~\ref{efficiency}. Patch-based approaches include our proposed PicT, WSPLIN, and IOPLIN. We report the efficiencies of two WSPLIN approaches, namely WSPLIN-IP (the default version) and WSPLIN-SS (the speed-up version). The backbone of PicT is Swin-S while the backbones of other patch-based approaches are EfficientNet-B3 (Effi-B3). From observations, we can see that Effi-B3 slightly performs better than Swin-S in efficiency. Swin-S needs 14\% more time for training the model with the same computing resources. Even so, PicT still shows significant advantages over other patch-based approaches in efficiency. For example, the training speed of PicT is around 7x, 8x, and 4x faster than the ones of WSPLIN-IP, IOPLIN, and WSPLIN-SS respectively. PicT can accomplish the label inferences of 26 more images, 14 more images, and 4 more images per second over WSPLIN-IP, IOPLIN, and WSPLIN-SS respectively. We attribute the good efficiency of PicT to the succinct learning framework of Swin-S. Besides the superiority in efficiency, PicT also holds a significant advantage in performance. For instance, WSPLIN-SS is the fastest patch-based approach except for PicT. PicT obtains 4.2\% and 6.1\% more performances than WSPLIN-SS in P@R=90\% and $F1$ respectively. To sum up, PicT is superior to the other patch-based approaches no matter in both performance and efficiency.

\begin{table}[tbp]
   \small
   \caption{ The training and inference efficiencies of different patch-based approaches. TrainTime is the total training time on one RTX3090 GPU. Throughput is measured on RTX3090 GPU FP32 with batch size 1.}
   \vspace{-0.3cm}
\begin{tabular}{p{1.9cm}p{1.1cm}<{\centering}p{1.1cm}<{\centering}rr}
       \toprule
       \textbf{Method} & \tabincell{c}{\textbf{I-DET}\\(P@R=90\%)} & \tabincell{c}{\textbf{I-REC} \\($F1$)}& \tabincell{c}{\textbf{TrainTime}\\(h)}& \tabincell{c}{\textbf{Throughput} \\(imgs/sec)}\\
       \midrule
       Effi-B3~\cite{effi}        & \underline{68.9\%}  & \underline{63.2\%}    & (+00\%)\hspace{5pt}0.7h                                          &(1.0x)~75\\
       Swin-S~\cite{swin}         & 79.9\%              & 64.9\%                & {\color{red}(+14\%)}\hspace{5pt}0.8h                        &{\color{red}(0.8x)}~62\\
       \midrule
       WSPLIN-IP~\cite{wsplin_trans}    & \underline{83.2\%}  & \underline{66.3\%}    &(+00\%)~11.1h                        & (1.0x)~31 \\
       IOPLIN~\cite{ioplin}       &  81.7\%             &  -                    &{\color{red}(+13\%)}\hspace{1pt}12.5h                       &{\color{blue}(1.4x)}~43 \\
       WSPLIN-SS~\cite{wsplin_trans}    &  81.4\%             &  64.1\%               &{\color{blue}(-49\%)}\hspace{5pt}5.7h                       & {\color{blue}(1.7x)}~53 \\
       PicT (ours)             & \textbf{85.6\%}     & \textbf{70.2\%}       & {\color{blue}(-87\%)}\hspace{5pt}1.5h                    &{\color{blue}(1.8x)}~57 \\
       \bottomrule
       \end{tabular}

   \label{efficiency}
   \vspace{-0.5cm}
\end{table}

\begin{table}[t]
  \small
  \caption{The ablation study results of the proposed model. The backbone is Swin-S, and PicT=\textit{Patch Labeling Teacher} + \textit{Patch Refiner}.}
  \vspace{-0.3cm}
\begin{tabular}{p{4.4cm}p{1.1cm}<{\centering}p{1.1cm}<{\centering}}
      \toprule
      \textbf{Method} & \tabincell{c}{\textbf{I-DET}\\(P@R=90\%)} & \tabincell{c}{\textbf{I-REC} \\($F1$)}\\
      \midrule
      Swin-S~\cite{swin} &  79.9\% &  64.9\% \\
      \midrule
      \textit{Patch Refiner} & 84.1\% & 67.0\% \\
      \textit{Patch Labeling Teacher} &  81.7\% &  67.3\%  \\
      \textit{Patch Labeling Teacher} + Swin-S & 85.1\% & 67.9\% \\
      \textit{Patch Labeling Teacher} + \textit{Patch Refiner} & 85.6\% & 70.2\% \\
      \bottomrule
      \end{tabular}
  \label{components}
   \vspace{-0.3cm}
\end{table}

\subsection{Ablation Study}

\begin{figure*}[t]
  \centering
  \includegraphics[width=16.5cm]{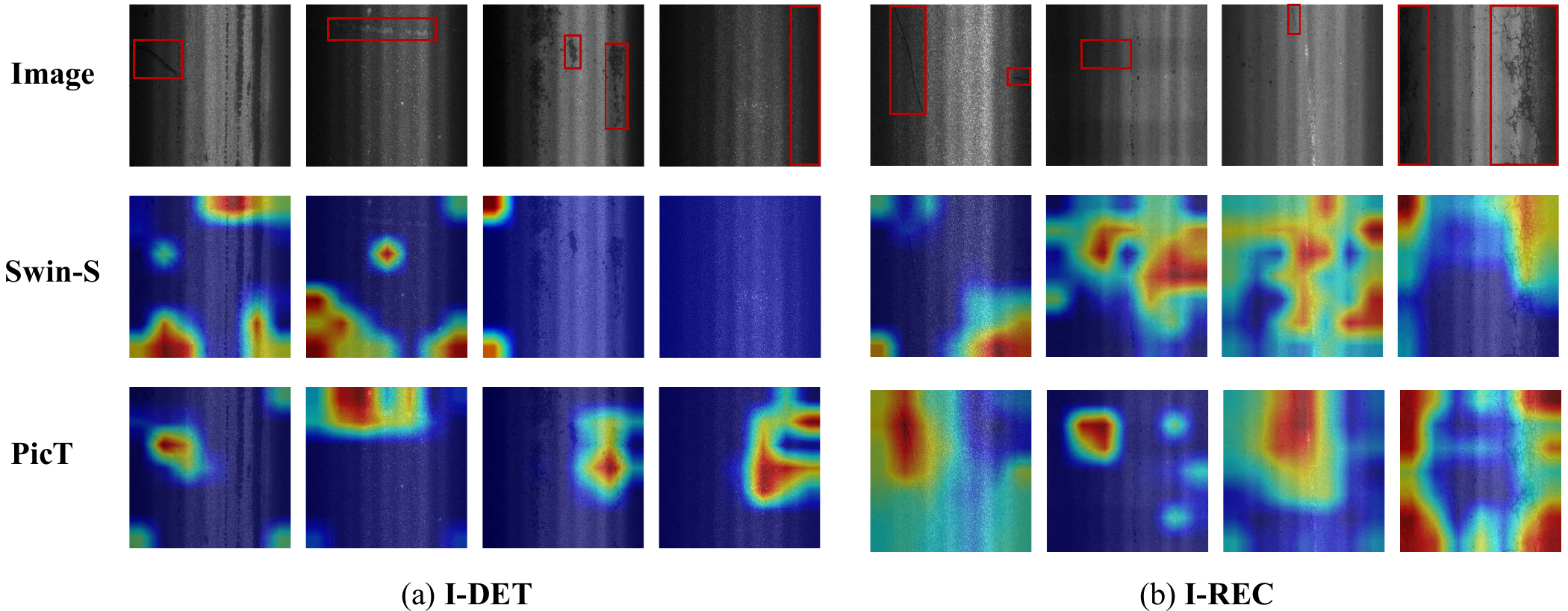}
  \vspace{-0.4cm}
  \caption{The CAM~\cite{cam} visualization of feature maps produced by PicT and Swin-S (The baseline). The first row is the raw images, and the red rectangle marks the distressed area. Best viewed in color.}
  \label{fig.cam}
  \vspace{-0.3cm}
\end{figure*}

\subsubsection{\textbf{Impacts of Different Modules}} 
Table~\ref{components} reports the performances of PicT under different settings in modules. Swin-S is the backbone of PicT, which is considered the baseline. \textit{Patch Refiner} and \textit{Patch Labeling Teacher} respectively indicate the PicT models only use our proposed image-level classification branch and our proposed patch-level classification branch. \textit{Patch Labeling Teacher} + Swin-S indicates the models use two classification branches. One is our proposed patch-level classification branch, and the other is the original image head of Swin-S (the broad head). \textit{Patch Labeling Teacher} + \textit{Patch Refiner} is the final PicT model, which has two classification branches. One is our proposed patch-level classification branch, and the other is our proposed image-level classification branch (the slim head). From observations, we can find that all two proposed modules improve the performances of the baseline no matter if used independently or in a combined way. These results verify our argument that further exploiting the discriminative information of patches via the patch-level supervision and highlighting the discriminative features of distressed patches can assist Swin-S in better addressing the PDC issue.

\begin{figure}[thb]
  \centering
  \includegraphics[width=8cm]{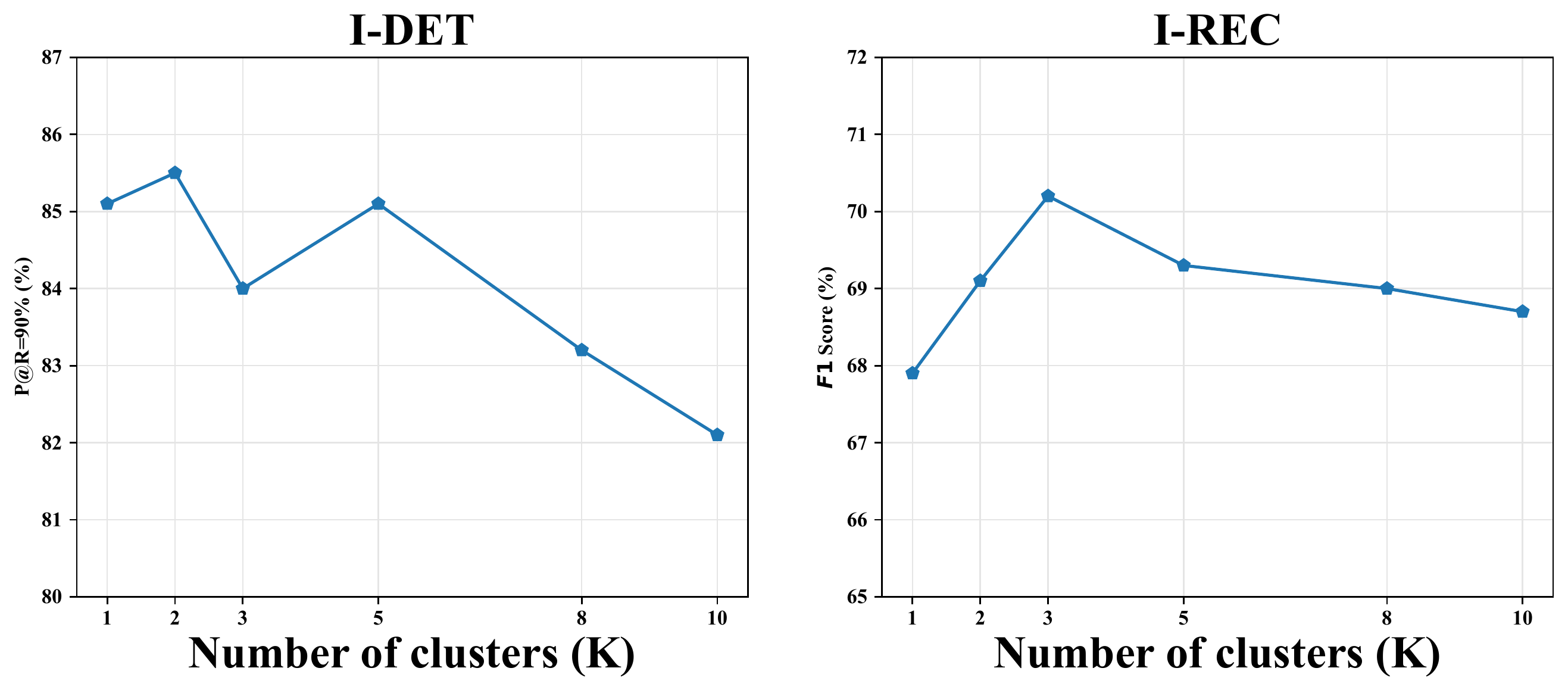}
 \vspace{-0.3cm}
  \caption{The performances of PicT under different $k$.}
  \label{fig.clu_k}
  \vspace{-0.3cm}
\end{figure}

\begin{figure}[thb]
  \centering
  \includegraphics[width=8cm]{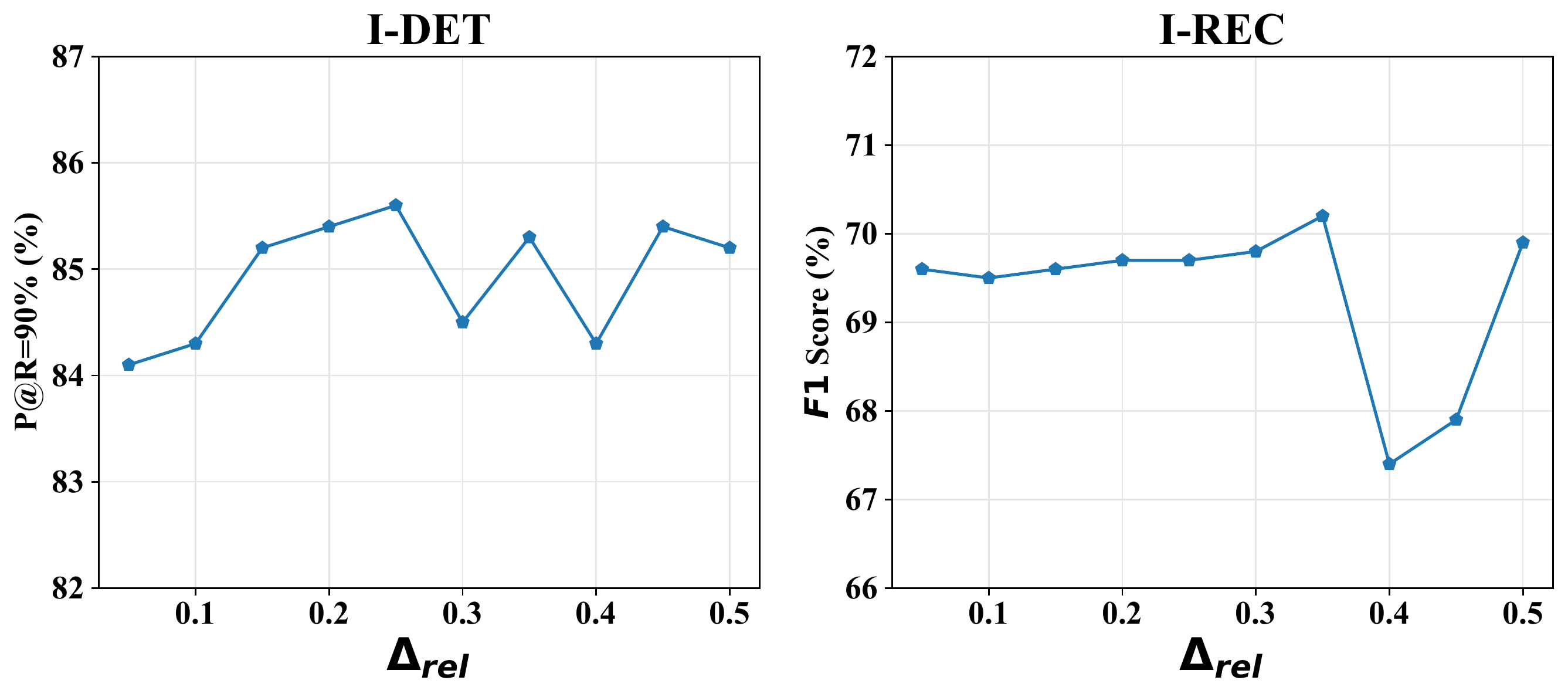}
   \vspace{-0.3cm}
  \caption{The performances of PicT under different $\Delta_{rel}$.}
  \label{fig.rdt}
  \vspace{-0.5cm}
\end{figure}

\subsubsection{\textbf{Discussion on Hyperparameters}} There are two tunable hyperparameters in our model. One is the cluster number $k$ in \textit{Patch Refiner}, which determines how many patches will be filtered out. The other is $\Delta_{rel}$ in the Relative Distress Threshold (RDT) of \textit{Patch Labeling Teacher}, which determines how many high-risk patches should be preserved in a distressed pavement image. Figures~\ref{fig.clu_k} and~\ref{fig.rdt} respectively plot the relationships between the performances of PicT and the values of these hyperparameters. We can observe that the PicT performs much better with a relatively smaller $k$. We attribute this to the fact that a larger $k$ is easy to cause the overmuch information loss in image-level classification, since only about $\frac{1}{k}$ of the patches have been preserved for taking part in the final classification. Similar phenomenon can be also observed on $\Delta_{rel}$. The reason why the relatively smaller $\Delta_{rel}$ often achieves a better performance is that the distressed area of pavement image is often very small and most of the patches are actually non-distressed, even in a distressed image. According to the observations of Figures~\ref{fig.clu_k} and~\ref{fig.rdt}, the optimal $k$ is 2 and 3,  while the optimal $\Delta_{rel}$ is 0.25 and 0.35 on detection and recognition tasks respectively.

\subsection{Visualization Analysis} 
In this section, we attempt to understand the proposed approach through visualizations.  All example images are from the CQU-BPDD test set.
\subsubsection{\textbf{CAM Visualization}} We leverage Grad-CAM~\cite{grad_cam} to plot the Class Activation Maps (CAM)~\cite{cam} of the image features extracted by the PicT and backbone Swin-S in Figure~\ref{fig.cam}. We see that Swin-S does not achieve the desired results on pavement images. On the one hand, Swin-S loses some discriminative information at the patch level, since patch-level supervised information has still not been sufficiently mined, such as the CAMs in the third and fifth columns. On the other hand, its broad head makes the model more susceptible to interference by the complex pavement environment (see the first column) or causes dilution of the discriminative regions (see the fourth column). In contrast, PicT demonstrates the fuller patch-level information utilization and stronger discriminating power over Swin-S with our proposed modules.


\begin{figure}[thb]
  \centering
  \subfigure{
  \includegraphics[width=3.5cm]{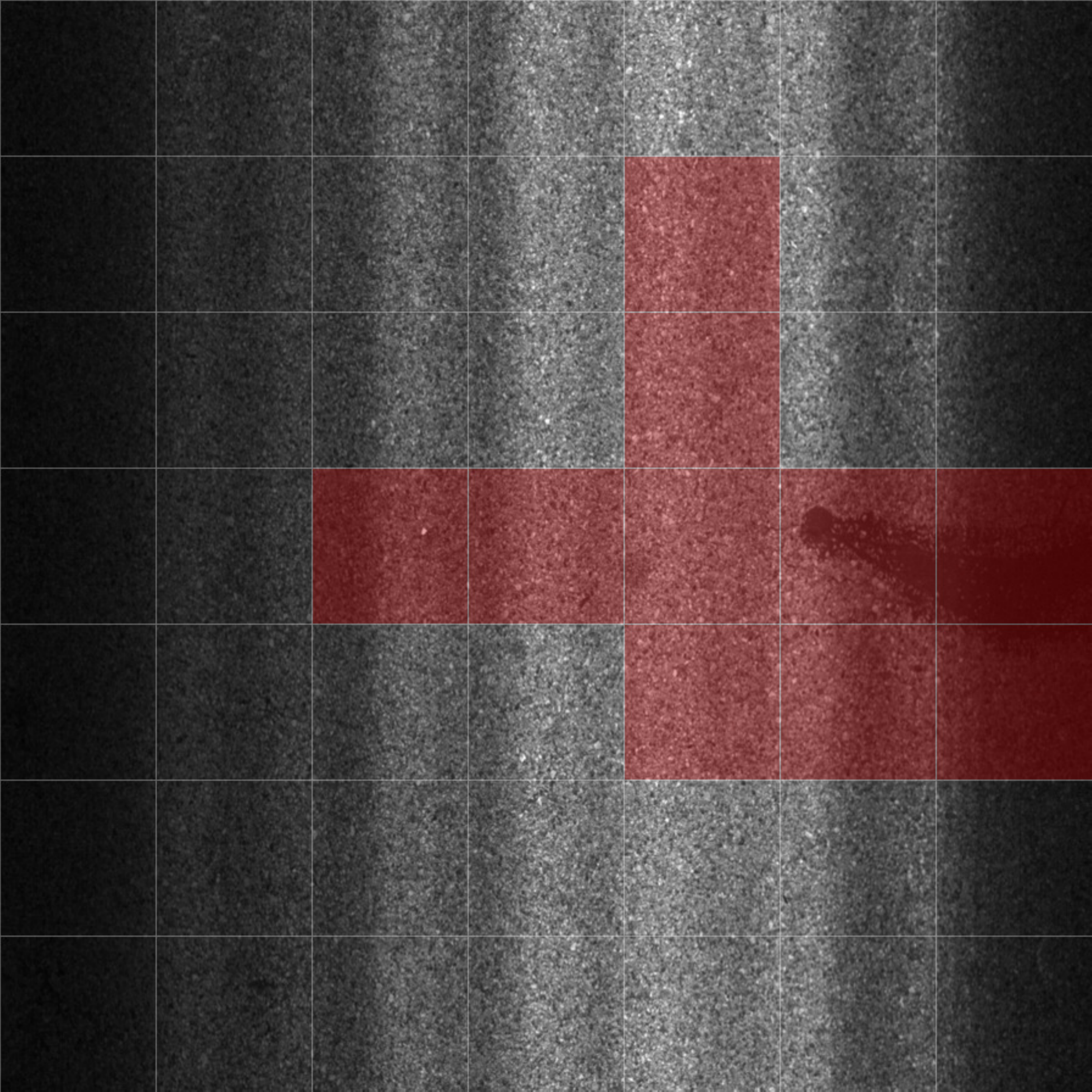}
  }
  \subfigure{
  \includegraphics[width=3.5cm]{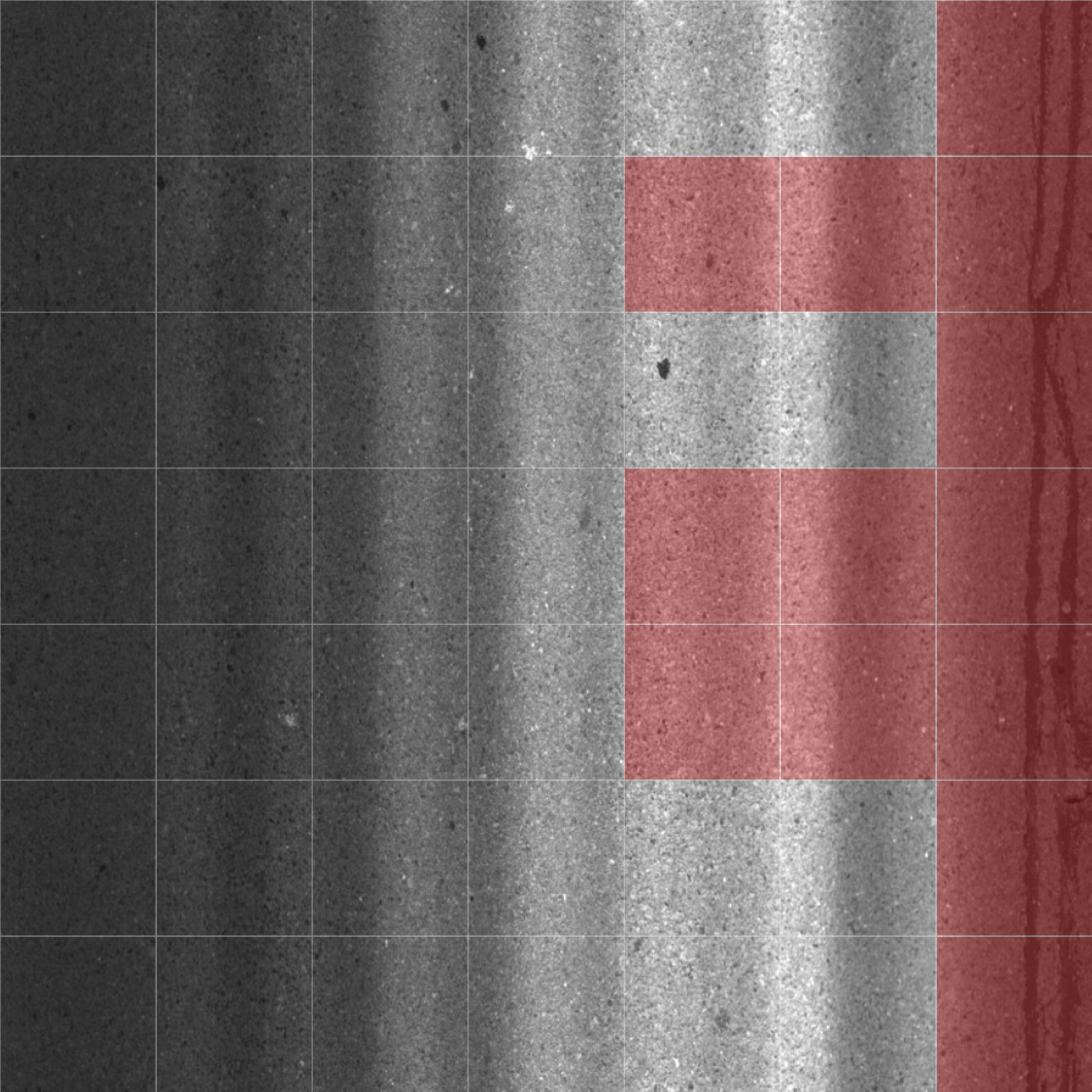}
  }
  \vspace{-0.3cm}
  \caption{The visualization of tokens based on the patch classification results produced by \textit{Patch Labeling Teacher}. The red mask indicates the distressed area predicted by the model.}
  \label{fig.token}
  \vspace{-0.8cm}
  \end{figure}
\subsubsection{\textbf{Token Visualization}} 
In Figure~\ref{fig.token}, we visualize the classification results of the tokens produced by \textit{Patch Labeling Teacher} to investigate whether the token can characterize a $16\times16$ pixel patch of pavement image. We can observe that these tokens exhibit strong local discriminability despite the absence of any strong supervised information. We attribute this to the local representational ability of the visual tokens themselves, and the effectiveness of the weakly supervised training module \textit{Patch Labeling Teacher}. Token visualization results demonstrate that the weakly supervised \textit{Patch Labeling Teacher} does indeed label the patches as designed. Moreover, it also explains how PicT profits from the patch-level classification branch to further improve the final classification performance.

\vspace{-0.2cm}
\section{Conclusion}
In this work, we propose a slim weakly supervised vision Transformer (PicT) for pavement distress classification, achieving solid results on detection and recognition tasks. We show that PicT elegantly solves the efficiency and modeling problems of previous approaches. We also present that PicT is more suitable for pavement distress classification tasks than the general vision Transformer due to its adequate and rational utilization of patches information with quantitative and visualization analysis.




\begin{acks}
  This work was supported in part by the National Natural Science Foundation of China under Grant 62176030, and the Natural Science Foundation of Chongqing under Grant cstc2021jcyj-msxmX0568.
\end{acks}

\bibliographystyle{ACM-Reference-Format}
\bibliography{acmart}

\end{document}